\documentclass[conference]{IEEEtran}
\pdfoutput=1
\IEEEoverridecommandlockouts
\usepackage{cite}
\usepackage{amsmath,amssymb,amsfonts}
\usepackage{graphicx}
\usepackage{textcomp}
\usepackage{xcolor}
\def\BibTeX{{\rm B\kern-.05em{\sc i\kern-.025em b}\kern-.08em
    T\kern-.1667em\lower.7ex\hbox{E}\kern-.125emX}}

\usepackage{changes}
\usepackage{comment}

\definechangesauthor[name="Ehsan", color=blue]{es}
\definechangesauthor[name="Korey", color=green]{kp}
\definechangesauthor[name="Ilya", color=red]{is}

\input{custom_commands.tex}
\input{algorithm_commands.tex}

\newcommand{\amlMlsvmFig}{0.8}

\newcommand{\amlResultFig}{.8}
\newcommand{\amlLiftUp}{-.9pc}
\newcommand{\amlHeaderUp}{.1pc}
\newcommand{\algFontSize}{\footnotesize}
\newcommand{\resFontSize}{\footnotesize}
\newcommand{\amlNotationTable}{\scriptsize}
\begin{document}

\setlength{\abovedisplayskip}{5pt}
\setlength{\belowdisplayskip}{5pt}
\bstctlcite{IEEEexample:BSTcontrol}

\title{AML-SVM: Adaptive Multilevel Learning with Support Vector Machines}

%
\author{\IEEEauthorblockN{1\textsuperscript{st} Ehsan Sadrfaridpour}
\IEEEauthorblockA{\textit{School of Computing} \\
\textit{Clemson University}\\
Clemson, USA \\
esadrfa@clemson.edu}
\and
\IEEEauthorblockN{2\textsuperscript{nd} Korey Palmer}
\IEEEauthorblockA{\textit{School of Computing} \\
\textit{Clemson University}\\
Clemson, USA \\
kpalmer@clemson.edu}
\and
\IEEEauthorblockN{3\textsuperscript{rd} Ilya Safro}
\IEEEauthorblockA{\textit{School of Computing} \\
\textit{Clemson University}\\
Clemson, USA \\
isafro@clemson.edu}
}

\maketitle

\begin{abstract}
    The support vector machines (SVM) is one of the most widely used and practical optimization based classification models in machine learning because of its interpretability and flexibility to produce high quality results. However, the big data imposes a certain difficulty to the most sophisticated but relatively slow versions of SVM, namely, the nonlinear SVM. 
The complexity of nonlinear SVM solvers and the number of elements in the kernel matrix quadratically increases with the number of samples in training data.  
Therefore, both runtime and memory requirements are negatively affected.
Moreover, the parameter fitting has extra kernel parameters to tune, which exacerbate the runtime even further. 
This paper proposes an adaptive multilevel learning framework for the nonlinear SVM, which addresses these challenges, improves the classification quality across the refinement process, and leverages multi-threaded parallel processing for better performance. 
The integration of parameter fitting in the hierarchical learning framework and adaptive process to stop unnecessary computation significantly reduce the running time while increase the overall performance. 
The experimental results demonstrate reduced variance on prediction over validation and test data across levels in the hierarchy, and significant speedup compared to state-of-the-art nonlinear SVM libraries without a decrease in the classification quality. The code is accessible at https://github.com/esadr/amlsvm.

\end{abstract}


\section{Introduction}\label{sec:introduction}
Support vector machine (SVM) is a widely used family of classification methods that leverage the principle of separating hyperplane. Technically, this is achieved by solving underlying regularized optimization model adjusting which can provide highly accurate and interpretable classification. 
Training linear SVM is very fast and can scale to millions of data points and features without using significant high-performance computing (HPC) resources. 
For problems that are not linearly separable, the nonlinear SVM uses the kernel trick by implicitly projecting the data into the higher-dimensional space to separate it by a hyperplane.
Nonlinear SVM usually reaches higher prediction quality on complex datasets. However, it comes with a price tag of being not scalable in comparison to its linear version.

Solving the Lagrangian dual problem is typically the way to cope with regularized nonlinear SVM models with the underlying convex quadratic programming (QP) problem. In a number of libraries (such as LibSVM  \cite{chang2011libsvm}) multiple methods have been implemented for solving both primal and dual problems. 
The complexity of the convex QP solvers for nonlinear SVM often scales  between $O(n^2f)$ to $O(n^3f)$~\cite{graf2004parallel} 
, where $n$ is the number of data points, and $f$ is the number of features.
Therefore, as $n$ increases, the running time of the solver also  increases, which hinders the usage of the nonlinear SVM for massive data sets. 


As the size of many datasets continues to grow due to the advancements in technologies such as high-throughput sequencing, e-commerce \cite{Heidari-etal-2020-Machinelearning} and the Internet of Things \cite{jagadish2015big}, more scalable machine learning algorithms are required. 
Therefore, while a nonlinear SVM is fast on small datasets and can provide highly accurate prediction, more research is required to develop scalable nonlinear SVM solvers for massive datasets. 

Our previous framework, Multilevel SVM (MLSVM), is a library that scales to millions of data points and exhibits up to two orders of magnitude faster running time compared to LIBSVM on reported benchmark  datasets~\cite{sadrfaridpour2019engineering}. 
The MLSVM leverages multilevel algorithms that are inspired by the algebraic multigrid and some of its restricted versions \cite{vlsicad}. These algorithms are known to be successful in accelerating computational optimization and modeling tasks without a loss in the quality. Examples include hypergraph partitioning \cite{shaydulin2019relaxation}, graph generation \cite{Gutfraind2012}, and clustering~\cite{d2019bootstrap}. 
Such multilevel methods have two essential phases, namely, coarsening and uncoarsening~\cite{vlsicad}. 
The coarsening phase gradually reduces the original problem size and generates a chain of smaller  problems that approximate the original one. This is done by constructing 
a (possibly fuzzy) hierarchy of aggregated data points. 
The uncoarsening phase starts from the smallest generated problem and gradually uses generated approximations to refine the solution of the original problem. During the uncoarsening phase, a refinement leverages a solution inherited from the previous coarser problem and is performed using a local processing solver to avoid any heavy computation.

The multilevel SVM \cite{sadrfaridpour2016algebraic,razzaghi2015scalable,sadrfaridpour2019engineering} is an approach to accelerate traditional SVM model solvers which also often improves the model quality as it exploits the geometry of data before starting the training. The key principles of multilevel SVM are 
as follows:\\
\noindent {\bf 1) Learning the separating hyperplane occurs within a small proximity to support vectors.} 
\noindent {\bf 2) Perform computationally expensive parts of the training only at the coarse levels.} 
\noindent {\bf 3) The best final model is not necessarily the finest one.}

We discussed several other principles in \cite{sadrfaridpour2019engineering}. Among them is the one that discusses the way multilevel SVM copes with imbalanced data.
While the MLSVM library \cite{sadrfaridpour2019engineering} provided fast runtime and highly accurate predictions on massive datasets, we observed several unexpected results on benchmark datasets that motivated us to extend that work and introduce the Adaptive MLSVM in this paper.

In multilevel algorithms \cite{safro:relaxml,vlsicad}, it is expected to observe that the quality of optimized objective is improved at each next finer level during the uncoarsening. As such, we expected to see a non-decreasing quality improvement across the uncoarsening. However, on some most difficult benchmarks, we observed that while the trained coarse models have exhibited reasonable to optimal prediction quality, some of the middle levels experienced a significant drop in the quality. In some cases, these middle level drops were continuing to the fine levels and in some were gradually improved back to the high quality results. 
Although the observed quality drops were rare, they often led to accumulating and increasing errors during the uncoarsening. Indeed, low quality support vectors may potentially be disaggregated into the data points that are even more distant to the optimal hyperplane which gives too much freedom to the optimizer. Other stages in the MLSVM pipeline such as the filtering of data points after too aggressive disaggregation were also affected.

\noindent {\bf Our Contribution} In this paper, we propose the Adaptive Multilevel SVM, a successful approach to detect the problem of inconsistent learning quality during the uncoarsening and efficiently mitigate it. At each level of the multilevel hierarchy, we detect the problem of quality decrease by validating the model using the finest level data.
In the adaptive multilevel SVM learning framework, we adjust the  training data by filling the training data gap with the misclassified validation data points, retraining the model and improving the quality with a new set of support vectors. 

Our exhaustive experimental results on the benchmark datasets demonstrate the proposed method recovers the multilevel framework from such quality drops and achieves higher quality in comparison to the non-adaptive  multilevel SVM as well as other state of the art solvers.
%
In addition, we speed up even more the runtime compared to non-adaptive multilevel SVM and reduce the prediction quality variance. 
Moreover, in the new version of the MLSVM library we implement the multi-thread support for parameter fitting to speed up the model training at each level. 
Our implementation is open-source and available at \url{http://github.com/esadr/amlsvm}.

\section{Related Works}
Improving the performance of SVM has been widely studied with a common goal of improving the training time without a significant decrease in prediction accuracy such as recent researches 
\cite{schlag2019faster_short,shalev2011pegasos,wen2018thundersvm,zhang2012maximum}. 


The complexity of nonlinear SVM on massive datasets is a potential target to develop scalable SVMs. 
A challenge with solving the quadratic programming problem for the kernel SVM is the kernel matrix of size $n^2$, where $n$ is the number of training data points. The matrix requires large memory, which is not feasible for massive datasets. 
The main categories of accelerated solvers follow one of the following approaches: a) target the solver performance directly, b) partition the original data, or c) considering alternative representation for the training data (including the multilevel approach), and their combinations. Some of the improvements also rely on the advancement of software infrastructure, hardware, and distributed frameworks.

\textbf{Solver performance:} 
A Sequential Minimal Optimization (SMO)\cite{cao2006parallel} has been used to solve the underlying quadratic programming problem. 
The ThunderSVM \cite{wen2018thundersvm} directly works on improving the performance of the solver using multiple cores and GPUs.

\textbf{Partition original training data:}
Instead of solving the QP with a large number of data points, a set of smaller problems can be generated and solved independently. This reduces the running time of training and memory utilization required to store the kernel matrix. A drawback to this approach is the quality of partitioning the original data and the quality of the approach to combine all the trained models to drive a final model. 
DC-SVM\cite{hsieh2014divide} used adaptive clustering for dividing the training data 
and relied on coordinate descent method to converge to a solution from multiple models. 
A disadvantage for these approaches is relying on the partitioning or clustering 
that is sensitive to the order of the training data, number of clusters. Assignment of a point to a partition or cluster is strict, which limits the points to move between clusters or participate in multiple clusters.

\textbf{Graph representation of training data:}
There are two schemas for partitioning the data. The earlier approaches rely on the original data in the feature space, while newer approaches such as\cite{razzaghi2015scalable,sadrfaridpour2019engineering,schlag2019faster_short} rely on a graph representation of data. The graph representation of the training data is constructed as a preprocessing step for partitioning which is explained in detail in section~\ref{sec:proximity_graphs}.
The graph representation provides the opportunity to leverage the multilevel paradigm that has been used successfully in a wide range of problems. We briefly mention some of the multilevel research related to graphs such as  graph partitioning~\cite{sanders2011engineering}, graph clustering~\cite{DhillonGuanKulis05fast}, and image segmentation~\cite{Grady06isoperimetricgraph}. 
%
The advantage of using a graph and multilevel paradigm such as Algebraic Multigrid (AMG) is to exploit more relaxed and less strict assignments of the data points to smaller aggregates compare to clustering, which has strict assignments and larger cluster sizes.

\textbf{The position of AML-SVM:}
Our proposed framework (AML-SVM) has a robust coarsening, which allows partial participation of points in multiple aggregates through gradual assignment and relaxation steps. Our results demonstrate that a small training data at the coarsest level can be used to train a model with high accuracy.
AML-SVM can directly leverage the performance improvements introduced by advanced solvers with multi-core or GPU support to achieve even faster runtime without any change in the coarsening or refinement process.
The uncoarsening (refinement) phase as a general step in multilevel methods has the potential for improvement of initial solutions and carrying essential information such as support vectors, and parameters to other levels. 
The improvements proposed in this paper reduce the variance between the quality of models at various levels and reduce the sensitivity of the framework to configuration parameters. 
Furthermore, these improvements, along with multiple core support for faster parameter fitting, can be used for ensemble models that we have not explored.

\section{Preliminaries}
Consider a set $\JJ$ of input samples that contains $n$ data points denoted by $x_i$, where $x_i \in \RR^d$, $1\leq i\leq n$. Each data point $x_i$ has a corresponding label $y_i \in \{-1, 1\}$. 
The SVM as a binary classification finds the largest margin hyperplane that separates the two classes of labeled data. The solution to the following optimization problem produces the largest margin hyperplane, which is defined by $w$, and $b$.

\algFontSize
\begin{align}\label{eqn:aml_SVM}
\textsf{minimize} & &  \frac{1}{2}\lVert w \rVert^2 + {C}\sum_{i=1}^n \xi_i \\
\textsf{subject to}  & & y_i (w^T \phi(x_i) + b) \geq 1 - \xi_i, &\quad i = 1, \dots,n \nonumber\\
& & \xi_i \geq 0, &\quad i=1, \dots, n \nonumber.
\end{align}
\normalsize
The misclassification is penalized using slack variables $\{\xi_i\}_{i=1}^n$.
The parameter $C > 0$ controls the magnitude of penalty for miscalssified data points.
The primal formulation in (\ref{eqn:aml_SVM}) is known as the \textit{soft margin} SVM  \cite{wu2005svm}.

The SVM takes advantage of kernel $\phi:\RR^d \rightarrow \RR^p ~ (d \leq p)$ to map data points to higher-dimensional space. The kernel measures the similarity for pairs of points $x_i$ and $x_j$. 
The Gaussian kernel (RBF), $\exp(-\gamma ||x_i-x_j||^2)$, which is known to be generally reliable  \cite{tay2001application}, is the default kernel in this paper.

For imbalanced datasets, different weights can be assigned to classes using the \emph{weighted} SVM (WSVM) to controllably penalize points in the minority class $C^+$ (e.g., rare events).
The set of slack variables is split into two disjoint sets $\{\xi_i^+\}_{i=1}^{n^+}$, and $\{\xi_i^-\}_{i=1}^{n^-}$, respectively.

In WSVM, the objective of (\ref{eqn:aml_SVM}) is changed into 

\algFontSize
\begin{align}\label{eqn:aml_WSVM2}
  \textsf{minimize} ~~~~~ & \frac{1}{2}\lVert w \rVert^2 + C\big( {W^+}\sum_{i=1}^{n^+}  \xi_i^+ + {W^-}\sum_{j=1}^{n^-}  \xi_j^- \big) .
\end{align}
\normalsize

The imbalanced data sets contain significanlty fewer positively labeled data points. Hence, the data points in the \emph{minority} class is denoted by $\JJ^+$, where size of minority class is $n^+ = |\JJ^+|$.
The rest of the points belongs to the \emph{majority} class which is denoted by $\JJ^-$, where $n^-=|\JJ^-|$, and $\JJ=\JJ^+ \cup \JJ^-$. 
We assign the class weight as the inverse of the total number of points in that class. For instance, the weight for minority class is $W^+= \frac{1}{n^+}$

In the our multilevel framework, one of the basic steps is aggregating data points. The aggregation can be performed either on full points or their fractions (i.e., a data point can be split into fractions, and different fractions can contribute themselves to different aggregates). Therefore, a coarse level data point normally contains several finer points or their fractions. Because of this, a data point can be more or less representative in comparison to other data points. We define and use the volume property for each point and calculate it as number of partial or fully participated original points in that aggregate.

We denote $v_i$ as volume for point $x_i$ which represent the internal importance of point $i$ among points in the same class. 
Moreover, we consider the class importance for the final weight of each data point. The calculation for both classes are the same, hence, we only present the positive class. We calculate the class weights based on sum of the volumes in the class, i.e.,
$W^+= \frac{1}{ \sum\limits_{i \in \JJ^+}^{} v_i }.$ 
We calculate the final weight for each point as product of its volume weight times class weight 
$\forall i\in \JJ^{+(-)} \quad W_i= W^{+(-)}\frac{v_i}{ \sum\limits_{j \in \JJ^{+(-)}}^{} v_j }.$
%
The instance weight based on volume and class, helps to improve prediction quality for the smaller class. 



\subsection{Two-level problem}
To describe the coarsening, uncoarsening and refinement algorithms, we introduce the two-level problem notation that can be extended into a full multilevel hierarchy (see Figure \ref{fig:aml_svm}). In Algorithm~\ref{alg:aml_general_refinement}, we will use subscript $(\cdot)_f$ and $(\cdot)_c$ to represent fine and coarse variables, respectively. For example, the data points of two consecutive levels, namely, fine and coarse, will be denoted by $\JJ_f$, and $\JJ_c$, respectively. The sets of fine and coarse support vectors are denoted by $\sv_f$, and $\sv_c$, respectively. We will also use a subscript in the parentheses to indicate the level number in the hierarchy where appropriate. For example, $\JJ_{(i)}$ will denote the set of data points at level $i$.

\subsection{Proximity graphs}
\label{sec:proximity_graphs}
The original MLSVM framework \cite{sadrfaridpour2019engineering} is based on the algebraic multigrid coarsening scheme for graphs  (see coarsening in \cite{SafroRB06graph,SafroRB08}) that creates a hierarchy of data proximity graphs. 
Initially, at the finest level, $\JJ$ is represented as two approximate $k$-nearest neighbor ($k$NN) graphs $G^+_{(0)} = (\GNode^+, E^+)$, and $G^-_{(0)} = (\GNode^-, E^-)$ for minority and majority classes, respectively, where each $x_i\in \GNode^{+(-)}$ corresponds to a node in $G^{+(-)}_{(0)}$. 

A pair of nodes in $G_{(0)}^{+(-)}$ is connected with an edge that belongs to $E^{+(-)}$ if one of these nodes belongs to a set of $k$-nearest neighbors of another. 
Throughout the multilevel hierarchies, in two-level representation, the fine and coarse level graphs will be denoted by $G^{+(-)}_f = (\GNode^{+(-)}_f, E^{+(-)}_f)$, and $G^{+(-)}_c = (\GNode^{+(-)}_c, E^{+(-)}_c)$, respectively. 
For $k$NN, our experiments show $k=10$, for all tested datasets, provides a good trade-off between computational cost of calculating $k$NN and average node degree for $G^{+(-)}$.

\section{Adaptive Multilevel SVM}
\begin{figure}
  \centering
  \includegraphics[width=\amlMlsvmFig\linewidth]{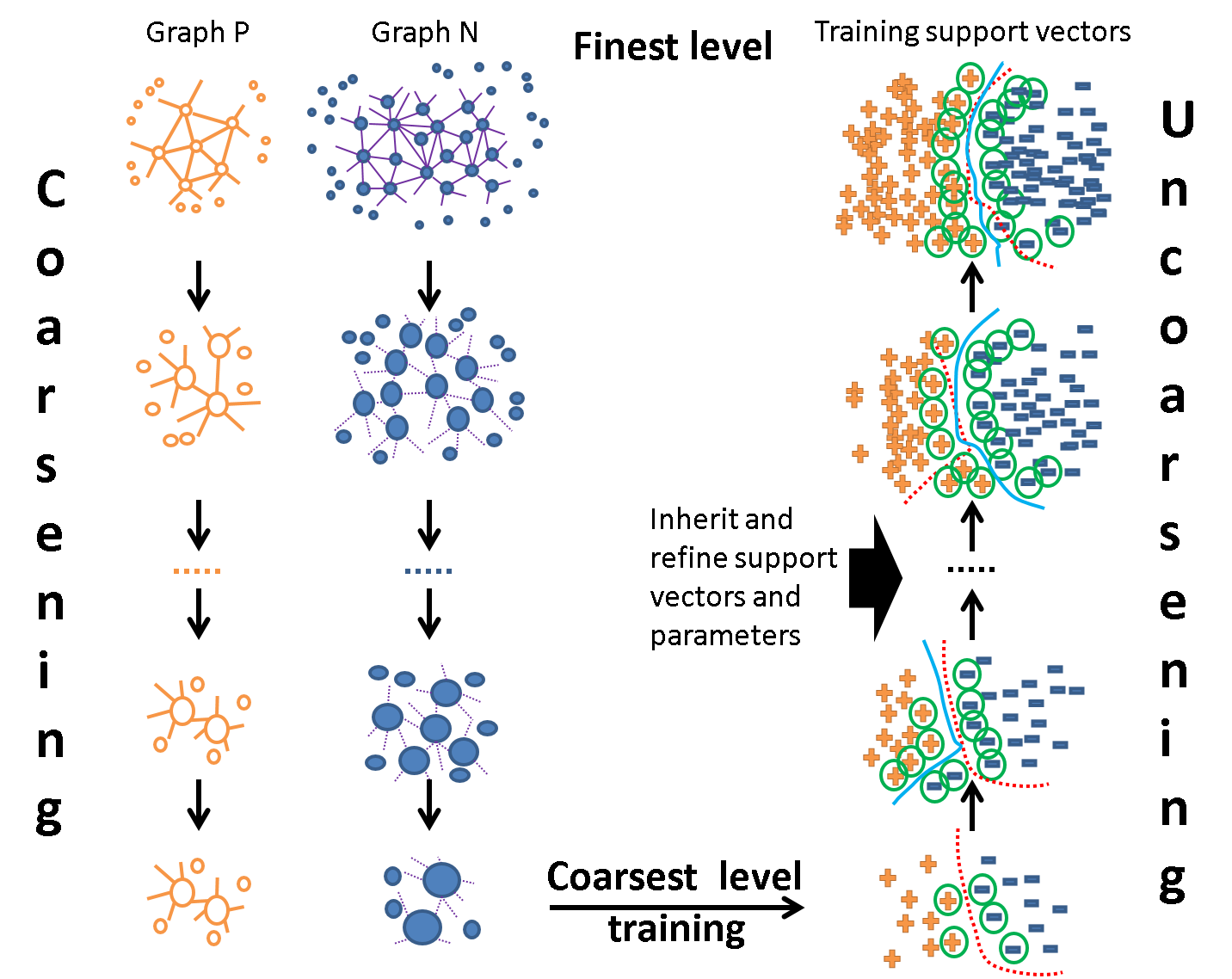}
    \caption{Adaptive multilevel learning framework scheme.}
    \label{fig:aml_svm}
\end{figure}

The multilevel support vector machine (MLSVM) ~\cite{sadrfaridpour2019engineering} 
includes three main phases (see Figure \ref{fig:aml_svm}), namely, (1) coarsening: gradual approximation of training set, (2) coarsest level solution: initial SVM training, and (3) uncoarsening: gradual support vector refinement at all levels. 
The coarsening phase creates a hierarchy of coarse training data $\JJ_{(i)}$, where $i$ is the level number where $i=1$ for the finest level and increases by one at each level of coarsening. The number of data points is decreasing at each level of coarsening, i.e.,  $|\JJ_{(i+1)}| < |\JJ_{(i)}|$. 
The coarsening continues until the aggregated training data reaches a certain threshold. The threshold depends on the available computational resources, which are capable of training a high-quality algorithm in a reasonable time. The threshold for the size of each class of data is denoted by $M$.
At the coarsest level $k$, the runtime to train a SVM model on a small training set is fast since training size is limited to less than a threshold for each class. ($\JJ_{(k)} < 2 \times M$). The output of trained SVM with the parameter fitting includes $\sv_{(k)}$ and the optimal parameters $C_{(k)}, \gamma_{(k)}$.
This output is inherited by the next finer level of the uncoarsening phase. At level $l$ of the uncoarsening, we select the next action among the three choices based on $\JJ_{(l)}$ and $Q_{(l)}$, which are the size of training data and the quality of model on validation data at level $l$, respectively:
\begin{itemize}
    \item Normal Refinement:
    The refinement continues by training new SVM model over training data based on decision boundary neighborhood at a coarser level. 
    \item Early Stopping: 
    We stop the refinement(uncoarsening) if the size of training data (both classes) is larger than threshold $\theta$ (line~\ref{ln:aml_amg_learning:while_size} in Algorithm~\ref{alg:aml_general_refinement}). 
    \item Detect and Recovery:
    If we detect a quality drop, we call the recovery algorithm described in Algorithm ~\ref{alg:augmentation}
\end{itemize}
The main parts of the framework are explained in Algorithms~\ref{alg:aml_amg_coarsening}, \ref{alg:aml_amg_coarsening_both_classes} and \ref{alg:aml_general_refinement}.

\scriptsize
\normalsize

The algebraic multigrid (AMG) SVM multilevel framework is inspired by the AMG aggregation solvers for computational optimization problems such as  \cite{Safro2012,safro-fastresp,SafroT11}. 
More information is covered in~\cite{sadrfaridpour2019engineering}.
The AMG coarsening provides high quality coarsening. It aggregates both fine points and fractions of points, which is more flexible than strict (or matching based)  coarsening techniques.
The fine level solution is identified using the interpolation operator on the coarse aggregates which are the main part of the coarsening phase in the AMG~\cite{mgbooktrott}.
%
The solution obtained from the coarsest level is gradually projected back to the finer levels using the interpolation and refined locally.

\subsection{Framework initialization} The AMG framework is initialized with $G^{+(-)}_0$ with the edge weights that represent the strength of connectivity between nodes to ``simulate'' the following interpolation scheme applied at the uncoarsening, in which strongly coupled nodes can interpolate solution to each other. 
In the classifier learning problems, this is expressed as a similarity measure between points. 
We define a distance function between nodes (or corresponding data points) as the inverse of the Euclidean distance in the feature space. 
For the completeness of the paper we need to repeat the coarsening phase we defined in \cite{sadrfaridpour2019engineering}.


\subsection{Coarsening Phase} 
In Algorithm~\ref{alg:aml_amg_coarsening}, we demonstrate the process of obtaining the training set (aggregated points) for the next coarser level. 
The process for both class are the same, hence, only minority class is explained. 
The input to the coarsening algorithm includes the fine level graph $G^+_f$, and data points (training set) $\JJ^+_f$, for the minority class.
The output includes the graph $G^+_c$ and data $\JJ^+_c$ at the coarser level for minority class. 
The same process applies to the majority class.

The process at level $f$ starts with selecting seeds. Seeds are nodes that serve as centers of aggregates at  level $c$. We introduce the notion of \emph{volume} $v_i\in \RR_+$ for all $i\in G^+_f$ that reflects the importance of each point during the coarsening phase. 
The volume is initialized to one for each node at the finest level. It increases during the coarsening with respect to the weighted sum of the volume of nodes that are aggregated together. (In some applications, so called anchor points can be used which can be reflected in the initial volumes.)
After the coarse graph is constructed, the aggregated data points are calculated based on interpolation matrix $P$. 

Seed selection starts with the future-volume $\vartheta_i$ computation which is a measure of how much an aggregate seeded by a node $i$ may potentially grow at the next level $c$, and it is computed in linear time:

\algFontSize
\begin{equation}
\label{eqn:aml_futurevolume}
\vartheta_i = v_i + {\sum\limits_{j\in \Gamma_i\cap \GNode^+_f} v_j \cdot \frac{w_{ji}}{\sum\limits_{k\in \Gamma_j \cap  \GNode^+_f} w_{jk}}},
\end{equation}
\normalsize
where $\Gamma_i$ is the neighborhood of node $i$ in $G^+_f$. 
We start with selecting a dominating set of seed nodes $S\subset \GNode^+_f$ to initialize aggregates. Nodes that are not selected to $S$ remain in $F$ such that $\GNode^+_f = F\cup S$. Initially, the set $F$ is set to be $\GNode^+_f$, and $S=\emptyset$  since no seeds have been selected. After that, points with $\vartheta_i > \eta \cdot \overline{\vartheta}$, i.e., those that are exceptionally larger than the average future volume are transferred to $S$ as the most ``representative'' points (line \ref{ln:aml_largevol}). 
Then, all points in $F$ are accessed in the decreasing order of $\vartheta_i$ updating $S$ iteratively (lines \ref{ln:aml_iterseeds1}-\ref{ln:aml_iterseeds2}), namely, if with the current $S$, and $F$, 
for point 
$i\in F$, $\sum_{j \in S} w_{ij} / \sum_{j \in \GNode^+_f} w_{ij}$
is less than or equal to some threshold $Q$.
%
%
The AMG interpolation matrix
$P\in \RR^{|G^+_f|\times |S|}$ is defined as

\algFontSize
\begin{equation} 
  \label{eq:aml_interp}
  P_{ij} =
  \left \{
  \begin{tabular}{cc}
    $ {w_{ij}} / {\sum\limits_{k \in \Gamma_i} w_{ik}} $ & if $i \in F$, $j \in \Gamma_i$ \\
    1 & if $i \in S$, $j=I(i)$\\
    0 & otherwise
  \end{tabular}
  \right \},
\end{equation}
\normalsize
where $\Gamma_i = \{j \in S \mid ij\in E^+_f\}$ is the set of $i$th seed neighbors, and $I(i)$ denotes the index of a coarse point at level $c$ that corresponds to the fine level aggregate seeded by $i\in S$. 

The aggregated points and volumes at a coarser level are calculated using the matrix $P$.
The edge between points $p=I(i)$ and $q=I(j)$ is assigned with weight

\algFontSize
\begin{equation}
\label{eq:aml_edgew}
w_{pq} = \sum\nolimits_{k \ne l} P_{ki} \cdot w_{kl} \cdot P_{lj}.
\end{equation}
\normalsize
The volume for the aggregate $I(i)$ in the coarse graph is computed by $\sum\nolimits_{j} v_j P_{ji}$.
i.e., the total volume of all points is preserved at all levels during the coarsening. The coarse point $q\in \GNode^+_c$ seeded by $i=I^{-1}(q)\in \GNode^+_f$ is represented by

\algFontSize
\begin{equation}\label{eq:aml_pointagg}
\sum_{j\in \mathcal{A}_i} P_{j,q} \cdot j,
\end{equation}
\normalsize
where $\mathcal{A}_i$ is a set of fine points in aggregate $i$. This set is extracted from the column of $P$ that corresponds to aggregate $i$ by considering rows $j$ with non-zero values.

\begin{algorithm}[H]
\algFontSize
\begin{algorithmic}[1]
  \State  $S\gets \emptyset,\quad \DD^+_f ,\quad F\gets \GNode^+_f$
  \State  {\textbf Calculate} using Eq. (\ref{eqn:aml_futurevolume}) $\forall i\in F$ $\vartheta_i$, and the average $\bar{\vartheta}$ \label{ln:aml_fvol}
  \State   $S \gets$ nodes with $\vartheta_i > \eta \cdot \overline{\vartheta}$ \label{ln:aml_largevol}
  \State   $F \gets \GNode^+_f \setminus S$ 
  \State   {\textbf Recompute} $\vartheta_i$ $\forall i \in F$
  \State   Sort $F$ in descending order of $\vartheta$ \label{ln:aml_iterseeds1}
  \State   \textbf{for} {$i \in F$}{
    \State   \quad \textbf{if} {$({\sum\limits_{j \in S} w_{ij}} / {\sum\limits_{j \in \JJ^+_f} w_{ij}}) \leq Q$ }{ \label{ln:aml_paramQ}
		   \State \quad \quad  move $i$ from $F$ to $S$
        } 
    }  
  \State \quad \textbf{end}    
  \State \textbf{end}    \label{ln:aml_iterseeds2} 
  \State   Build interpolation matrix $P$ according to Eq. (\ref{eq:aml_interp})
  \State   Build coarse graph $G^+_c$ with edge weights using Eq. (\ref{eq:aml_edgew})
  \State   Define volumes of coarse data points 
  \State   Compute coarse data points $\DD_c^+$ 
   \State  \textbf{return}    $(\DD_c^+, G_c^+)$
\caption{AMG coarsening for one class}
\label{alg:aml_amg_coarsening}
\end{algorithmic}
\end{algorithm}
\normalsize
%
%
The complete coarsening phase for both classes are demonstrated in Algorithm~\ref{alg:aml_amg_coarsening_both_classes}.  
The parameter $M$ controls the size of each class at the coarsest level. In our experiments, the $M$ is set to 300 which ensures training a SVM model using LIBSVM is fast.




\begin{algorithm}[H]
\algFontSize
\begin{algorithmic}[1]
\caption{Coarsening phase for both classes}
\label{alg:aml_amg_coarsening_both_classes}
    \State \textbf{if} {$|\GNode^+_f| \leq M \And |\GNode^-_f| \leq M$} \textbf{then}  \Comment{Solve exact problem}
    \State \quad $(\sv_f, C^+, C^-, \gamma) \leftarrow $ Train SVM model on $\JJ_f$ (+NUD) \label{ln:aml_cst_train_model}  
    \State \textbf{else }  \Comment{Start or continue to coarsen the problem}
    \State \quad \textbf{if} {$|\GNode^+_f| \leq M$} \textbf{then} 
    \State \quad \quad $\DD^+_c \leftarrow \DD^+_f;~ G^+_c \leftarrow G^+_f$
    \State \quad \textbf{else}
    \State \quad \quad $(\DD^+_c, G^+_c) \leftarrow$ coarse $(\DD^+_f, G^+_f)$
    \State \quad \textbf{if} {$|\GNode^-_f| \leq M$} \textbf{then} 
    \State \quad \quad $\DD^-_c \leftarrow \DD^-_f;~ G^-_c \leftarrow G^-_f$
    \State \quad \textbf{else} 
    \State \quad \quad $(\DD^-_c, G^-_c) \leftarrow$ coarse $(\DD^-_f, G^-_f)$

\end{algorithmic}
\end{algorithm}
\normalsize

\subsection{Uncoarsening}
The primary goal of uncoarsening is to interpolate and refine the solution of coarse level $c$ for the current fine level $f$. One of the advantages of multilevel framework is limiting the size of the projected information. We only inherit the optimal parameters $C, \gamma$ and $\sv_c$. The corresponding aggregates to $\sv_c$ are added to $\JJ_f$, namely,
The training data at the finer level ($\JJ_f$) starts as an empty set, then for any aggregate $p$ ($\mathcal{A}_p$) which is part of support vectors at the coarser level ($\sv_c$), all points $j$ in aggregates are added to the training data.

\algFontSize
\begin{align}\label{eq:aml_amg-disagg} 
 \JJ_f \gets \emptyset; ~ &\quad  \forall p\in \sv_c \quad \forall j\in \mathcal{A}_p \quad \JJ_f \gets \JJ_f \cup j.
\end{align}
\normalsize
%
%

\begin{algorithm}
\algFontSize
\begin{algorithmic}[1]
\caption{Uncoarsening (refinement) phase} 
\label{alg:aml_general_refinement}
    \State Input: $(\sv_c, \tilde{C}^+, \tilde{C}^-, \tilde{\gamma})$ \Comment{Solution from a coarser level}
    \State Create training set by Eq. \eqref{eq:aml_amg-disagg}
    \State {While {$(|\DD^+| +|\DD^-|) <
    \theta$} 
    then}\label{ln:aml_amg_learning:while_size}
    \State \quad {Train SVM Models} \label{ln:aml_train_model2}
    \State \quad {Evaluate performance qualities} 
    \State \quad \textbf{Call detect and recovery Algorithm} \label{ln:aml_call_detect_recovery}
    \State \quad {Select best model based on G-mean, SN, and nSV} \label{ln:aml_sort}
    \quad \NoNumber{\Comment{Refine the solution}}
    \State \quad $\{(\sv_f, C^+, C^-, \gamma)_i\}_{i=1}^k \leftarrow$  refine$( \sv_f, \tilde{C}^+, \tilde{C}^-, \tilde{\gamma})$ \label{ln:aml_send_solution_to_finer_level}
%
    \State \textbf{return} best model from $k$ levels $\{(\sv_f, C^+, C^-, \gamma)_i\}_{i=1}^k$ \label{ln:aml_ret_best_model}
    
    \end{algorithmic}
\end{algorithm}
\normalsize
Algorithm \ref{alg:aml_general_refinement} demonstrate the overall process for the refinement process. 
\section{Learning Approach}
\subsection{Imbalanced classification} 
In addition to the cost-sensitive and weighted models solved in the refinement, the multilevel learning framework can cope with the imbalanced data and mitigates its negative effect on the classification quality for the minority class. 
We make a balanced training set at the coarsest level by performing the coarsening for each class independently until the number of data points is smaller than $M$.
While the minority class with a small number of points reaches this threshold in a few levels of coarsening, the majority class may take many more steps. 
The data for the minority class is preserved and transferred entirely to the coarser level. Therefore, at the coarsest level, the size of both classes are balanced.
The importance of points and imbalanced size is enforced through the volume for each data point.

\subsection{Parameter Fitting}
We use the Nested Uniform Design (NUD)~\cite{huang2007model} 
in our computational experiments for parameter fitting. 
In NUD, at first stage, multiple points are selected in the search space and models trained based on the parameters representing each of those point. In the following stages, the best point based on highest classification quality is selected as the center for the search. Multiple points around this are evaluated and the best results from all stages is the optimal parameters. 
In a similar approach, we pass the best parameters from a coarser level as the center for the first stage of NUD. Therefore, the inherited parameters from a coarser level stage is evaluated on the finer data.

\subsection{Learning Challenge}
\begin{figure}[bt]
\vspace{-10pt}
  \centering
  \hspace{-.5cm}\includegraphics[width=1.05\linewidth]{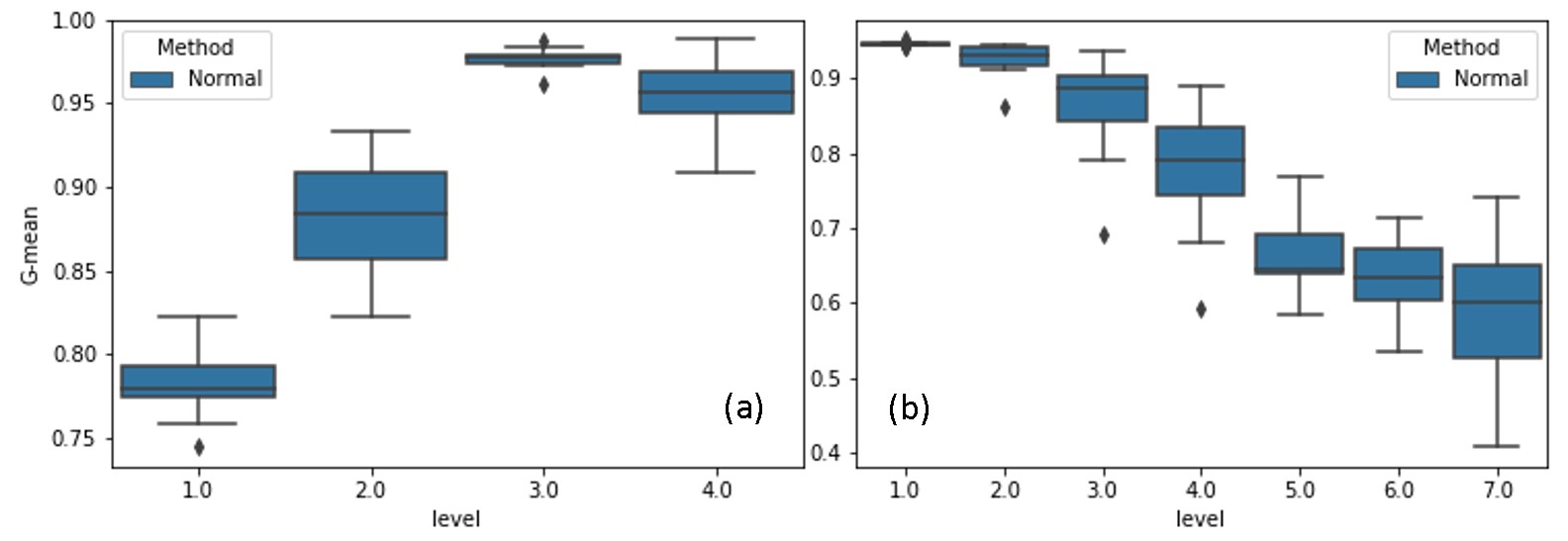}
  \vspace{-0.4cm}
    \caption{Natural (a) and unnatural (b) trends, increasing and decreasing the classification quality (G-mean), respectively, during refinement.}
    \label{fig:natural_trend}
\end{figure}

In the hierarchical framework, it is expected that the first solution is only a reasonable (but not necessarily the best)  approximation of the optimal solution. Therefore, extra steps are required to improve coarse solutions gradually. 
In the context of the classification task, the initial solution should provide a reasonable decision boundary. We can measure the quality of the initial solution with performance measures on validation data, which we cover in section~\ref{sec:experimental_results}. We can express it more accurately by
$Q_{(1)} \leq Q_{(2)} ...  \leq Q_{(k)},$
when $Q_{(i)}$ is the classification quality at level $i$. 
We denote this expected trend with a continuous increase in quality as the \textbf{natural trend} and Figure~\ref{fig:natural_trend}(a) demonstrates it for the Ringnorm data set. 
\scriptsize
\normalsize

However, this is not what we observed over exhaustive experiments with various data sets. In our experimental results, there were some cases in which the classification quality was dropped significantly during the refinement. The consequences of unexpected phenomena are as follows.

The training process for the models at the following levels of refinement will be based on sub-optimal training data. 
As explained earlier, the training data at each level of refinement is filtered to contain only the neighbor points of SVs of a model at the coarser level. When the SVs are sub-optimal, the filtering adds the points which are not representative of the data for the SVM to learn an optimal decision boundary. 
Moreover, the number of SVs may increase dramatically without significant improvement in the classification quality. The increase in the number of SVs drastically increases the number of SV's neighbors, which forms a larger training data at the next level. Therefore, it reduces the performance of the rest of levels in the refinement. 

We recognized a pattern that the quality improves during refinement but at some levels, declines significantly. We denote it as \textbf{unnatural trend} and Figure~\ref{fig:natural_trend}(b) demonstrates an example of the unnatural trend for the Cod-RNA dataset.

%
%

%
Benefits of detecting and addressing the decrease in quality during the refinement are:
\begin{itemize}
    \item Improving the classification quality of the finer levels and in general
    \item Improving the computational performance of the refinement process by improving the quality of training data set for the finer levels.
    \item Reducing the computational cost of refinement through the Early Stopping process when there is not a significant improvement in quality by training larger models
    \item Reducing the variance for the classification quality across levels
    \item Preserve smaller training set during the refinement
\end{itemize}

Intuitively, the coarsening procedure gradually creates approximated (or summarized) representations of the finest level (original) data. An initial model we train at the coarsest level, provides a large margin hyperplane which can be used as the \emph{final solution}. 
On the one hand, at the finest level, rich data can easily lead to over-fitted models, a phenomenon frequently observed in practice \cite{dietterich1995overfitting}. On the other hand, over-compressed data representation may lead to an under-fitted model due to lack of detail information. 
Our experiments confirm that more than half of the best models are obtained from the coarse (but not coarsest) and middle levels which typically prevents over- and under-fitting.
In general, if the best models were produced by the  finest and middle levels, we recommend to use the model obtained at the middle levels to avoid potential over-fitting. This recommendation is based on the observation that same quality models can be generated by different hyperplanes but finest models may contain a large number of support vectors that can lead to over-fitting and slower prediction. However, this is a hypothesis that requires further exploration. In our experiments, no additional parameters or conditions are introduced to choose the final model. We simply choose the best model among those generated at different levels.

\label{sec:proposed_method}
\subsection{Proposed Adaptive Method to Recover the Quality Drop}
The drop in classification quality is equivalent to an increase in the number of points that are misclassified. Therefore, a classifier quality can be improved by training on more similar data to misclassified data points. The training data at each level is a sub-sample of data around the decision boundary (hyperplane) at the coarser level. The data is sampled from the neighborhood of SVs from the trained model at the coarser level. The points which are misclassified in the training data are not crucial for improvement since the SVM did not find a better solution for the decision boundary. However, the validation data points which are misclassified are the crucial points that need to be classified correctly in this step. It is worth to notice, this step does not rely on any information from the test data and only rely on the validation and training data. 
Increasing the number of data points usually help the classifier to learn a better decision boundary. A $p$,$n$ number of neighbors of misclassified validation data points are found in the data from positive and negative class, respectively. The neighbor points are added to the training data, and the model is retrained. The $p$, $n$ can start from 1 and increase to larger values. However, as these parameters increase, the number of data points that are added to training data increase with respect to the validation data set size and affect the computational performance in the current level and overall framework. 
We mention important considerations of earlier work in the MLSVM framework.

It is essential to mention that for the validation data, we used $FF$ method in MLSVM that is based on the sampling of the original training data at the finest level. 
Therefore, the size of the validation data set for large data sets is significantly larger than the training data at the coarser levels. For instance, a data set with $5M$ data points such as SUSY with k-fold cross-validation using $k=5$ has $4M$ training data at the finest level. Using the validation sample rate of $10\%$, the validation size at all levels has $400k$ data points. At the coarsest level, the training size is less than $2 \times M$ (600 samples with default $M=300$)
, and it continues to grow slowly over the next few levels of refinement. Suppose the accuracy is $80\%$ at a level. The maximum number of points that can be added to training data would be $400k* 20\% *(p+n)$ or $80k * (p+n)$. This would be an impractical training size. Therefore, we limit the number of $p$ and $n$ to one. In our experiments, many miclassified points have common neighbors which were not in the training. We only add distinct points into training set, hence, the increase is training size is much smaller.
    %
    
The focus of MLSVM is on extremely imbalanced data sets. 
%
The majority of data points belong to the larger class in highly imbalanced data sets. Therefore, a more restrict sampling ratio helps to reduce the number of data points from the majority or larger class in the validation data. However, the minority or smaller class typically has ten times fewer data points, and its size would not cause performance problems. Therefore, a more significant sample rate, such as $50\%$ or more for minority class compares to $10\%$ or less for majority class, both reduce the number of points in the validation data and make the validation data better representative sample to evaluate the quality of the classifier. A validation set is significant for the framework, which allows us to find the best model from many models at various levels. If the validation data is not representative, the quality of the model will severely affect the test data, which is the ultimate goal of training a classifier.
    The validation data size for a large data set with a sample ratio of $10\%$ is still huge. Using smaller values such as $2\%$, the validation data was not representative of test data anymore in the past. With proposed new unbalanced sampling ratios for validation, the $2\%$ validation sample ratio of majority class, and $50\%$ sample ratio of the minority class, reduce the computational challenge.
The notation used in Algorithm~\ref{alg:augmentation} is  explained in Table~\ref{tbl:augmention_notation}.
\vspace{-.2cm}
\begin{table}[b]
\vspace{-.5cm}
\caption{Notations in Algorithm~\ref{alg:augmentation}}
\amlNotationTable
\begin{tabular}{c|l}
\hline
    Notation    & Description \\ \hline
    $k$         & Coarsest level identifier     \\ 
    $c$         & Current level identifier      \\ 
    $Q$         & Predictive Quality metric     \\
    $\delta$    & Parameter: Significant quality threshold  \\
    $\JJ$       & Only neighbors of $\sv$ in the training set   \\
    ${\JJ}_{all}$ & Complete training set at current level   \\
    ${\JJ}^{a}$     & Augmented $\JJ$ (neighbors of $\sv$)      \\
    $A_i$       & Augmented point which is a NN of misclassified point $i$ in $\JJ$ \\
    $\VV$       & Validation set               \\ 
    $\FP$       & Set of False Positive predictions in validation set  \\
    $\FN$       & Set of False Negative predictions in validation set   \\    
    $NN^{+(-)}_i$    & Nearest Neighbors in ${\JJ}^{+(-)}_{all}$ for data point $i$  \\
\end{tabular}
\label{tbl:augmention_notation}
\end{table}
\normalsize
%

\begin{algorithm}[H]
\algFontSize
\caption{Detection and recovery of quality drop.}
\label{alg:augmentation}
\begin{algorithmic}[1]
\small
\State $Q_{(max)} = \max \ Q_{(i)}$  \\
        \quad  \quad  \quad  \quad {$\text{where} \  i \in \{k, k-1, k-2,..., c-1\}$}   \label{ln:alg_aug:detect_drop}
\State {Evaluate performance qualities for all models (NUD)}
\State {$Q_{(c)} \gets $ G-mean for the best model at current level}
\State {if $Q_{(c)} > Q_{(max)}$;} {then}
\State \quad $Q_{(max)} = Q_{(c)}$
\State {else }
\State \quad {if $Q_{(max)} - Q_{(c)} > \delta$;} {then}
\State \quad \quad $\FP \gets $ find false positive points in the validation set
\State \quad \quad $\FN \gets $ find false negative points in the validation set
\State \quad \quad {for $i \in \{\FP \cup \FN\}$;}  {then}
\State \quad \quad \quad ${A_i} \gets NN^{+}_i \cup NN^-_i$ \Comment{new points with respect to $i$}

\State \quad \quad ${\JJ^{a}} \gets \JJ \cup A$    
\State \quad \quad {Train new SVM models on $\JJ^a$ (NUD)}       \label{alg_aug_ln:retrain_model}  
\State \quad \quad {Evaluate performance qualities for all new models}
\State \quad \quad {$Q^a_{(c)} \gets $ G-mean for the best of new models}

\State \quad {if $Q^a_{(c)} > Q_{(c)}$;} {then} 
\State \quad \quad {Return}{$\{\sv^a_c, C^a, \gamma^a \}$} \Comment{Improved the quality}
\State {Return}{$\{\sv_c, C, \gamma \}$} \Comment{No improvement}
\end{algorithmic}
\end{algorithm}
\vspace{-.5cm}
\normalsize

\normalsize
\section{Experimental Results}
\label{sec:experimental_results}
First, we report the comparison between the ML-SVM algorithm and the state-of-the-art SVM algorithms such as LIBSVM and DC-SVM in terms of classification quality and computational performance. Then, we compare the proposed AML-SVM method and ML-SVM method using new set of experiments with more details.
\subsection{Performance measures}
Performance measures are metrics which are used to evaluate the classification quality of the model. We report the 
accuracy (ACC), recall or sensitivity (SN), specificity (SP), and geometric mean (G-mean). They defined as 
\algFontSize
${\textrm{ACC}=\frac{TP+TN}{FP+TN+TP+FN}}, {\textrm{SN}=\frac{TP}{TP+FN}}, 
 {\textrm{SP}=\frac{TN}{TN+FP}},$ 
 ${\textrm{G-mean}=\sqrt{\textrm{SP}\cdot \textrm{SN}}},$
\normalsize
Where $TN$, $TP$, $FP$, and $FN$ correspond to the numbers of real negative, true positive, false positive, and false negative points. 
Our primary metric for comparison is G-mean, which measures the balance between classification quality on both the majority and minority classes. This metric is illuminating for imbalanced classification as a low G-mean is an indication of low-quality classification of the positive data points even if the negative points classification is of high quality. This measure indicates the over-fitting of the negative class and under-fitting of the positive class, a critical problem in imbalanced data sets. 
In both ML and AML -SVM frameworks, many models are trained. We need to provide one value for prediction as to the final model performance measure. Therefore, we evaluate all the models using the validation set and select the best model and report the performance measures of the selected model over the test (hold-out) set in the results.
The detail of imbalanced data sets used in our experiment is presented in Table \ref{tbl:aml_dataset_info}. The imbalance ratio of data sets is denoted by $\epsilon$.

\begin{table}[bt]
\caption{Benchmark data sets.}
\vspace{-.1cm}
\resFontSize
\centering
    \begin{tabular}{lccccccc}         
	\hline
		Dataset	&	$\epsilon$	&	$n_f$ &	$|\JJ|$	&	$|\CC^+|$	&	$|\CC^-|$	\\ \hline
		Advertisement	  &	0.86	&	1558    &	3279	&	459	    &	2820	\\
		Buzz              &	0.80	&	77	    &	140707	&	27775	&	112932	\\
		Clean (Musk)	  &	0.85	&	166	    &	6598	&	1017	&	5581	\\
		Cod-rna	          &	0.67	&	8	    &	59535	&	19845	&	39690	\\
		Forest (Class 5)  & 0.98    &   54      &   581012  &   9493    &   571519  \\
		Letter  	      &	0.96	&	16	    &	20000	&	734	    &	19266	\\
		Nursery	          &	0.67	&	8	    &	12960	&	4320	&	8640	\\
		Ringnorm	      &	0.50	&	20	    &	7400	&	3664	&	3736	\\
		Twonorm	          &	0.50	&	20	    &	7400	&	3703	&	3697	\\
		HIGGS             & 0.53    & 28        & 11000000  & 5170877   & 5829123    \\
		SUSY              & 0.54    & 18        & 5000000   & 2287827   & 2712173    \\	\hline
	\end{tabular}

	\label{tbl:aml_dataset_info}
 	\vspace{-.3cm}
\end{table}
\normalsize

\subsection{LIBSVM, DC-SVM and ML-SVM with AML-SVM Comparison}
For self completeness of this paper and convenience of the readers, we present all the relevant results from the ML-SVM paper. The following results provides the clear comparison between the state-of-the-art methods which set the base for next section which we provide more detailed results for comparing the ML-SVM and AML-SVM methods. 

\begin{table}[bt]
\caption{Performance measures and time for LIBSVM, DC-SVM, ML-SVM and AML-SVM on medium size benchmark data sets.}
\vspace{-.5cm}
\scriptsize
\setlength{\tabcolsep}{1.1pt}
  \begin{center}
    \scalebox{.95}{
    \begin{tabular}{lcc|cc|cc|ccc}
      \hline
       & \multicolumn{2}{c|}{LIBSVM} & \multicolumn{2}{c|}{DC-SVM} & \multicolumn{2}{c|}{ML-SVM} 
       & \multicolumn{3}{c}{AML-SVM} \\ \cline{2-10}
            Datasets        & G-mean     &C-Time     & G-mean     & C-Time     & G-mean     & C-Time       & G-mean    & C-Time    &W-Time \\ \hline
            Advertisement   & 0.67       & 231      & 0.90       & 610      & \bf{0.91}  & 213        & 0.90      & \bf{126} & \bf{23.84  }   \\
            Buzz            & 0.89       & 26026    & 0.92       & 2524     & \bf{0.95}  & \bf{31 }   & \bf{0.95} & 269      & \bf{64.13  }   \\
            Clean (Musk)    & \bf{0.99}  & 82       & 0.94       & 95       & \bf{0.99}  & 94         & 0.93      & \bf{21}  & \bf{4.32   }   \\
            Cod-RNA         & \bf{0.96}  & 1857     & 0.93       & 420      & 0.94       & \bf{13  }  & 0.95      & 75       & \bf{16.83  }   \\
            Forest          & 0.92       & 353210   & \bf{0.94}  & 19970    & 0.88       & \bf{948}   & 0.88      & 2149     & \bf{394.97}    \\
            Letter          & 0.99       & 139      & \bf{1.00}  & 38       & 0.99       & 30         & 0.98      & \bf{14}  & \bf{3.04   }   \\
            Ringnorm        & \bf{0.98}  & 26       & 0.95       & 38       & \bf{0.98}  & \bf{2 }    & 0.97      & 3        & \bf{0.77   }   \\
            Twonorm         & \bf{0.98}  & 28       & 0.97       & 30       & \bf{0.98}  & \bf{1 }    & 0.97      & \bf{1}   & \bf{0.42   }   \\ \hline
    \end{tabular}
    }
  \end{center}
  \label{tab:aml_mlsvm_journal_paper_medium_data_time_performance_measures}
  \vspace{-.5cm}
\end{table}

\begin{table}[!ht]
\caption{Computational time in seconds for LIBLINEAR, LIBSVM/DC-SVM, ML-SVM and AML-SVM on larger benchmark data sets}
\scriptsize
\setlength{\tabcolsep}{1.1pt}
  \begin{center}
    \scalebox{.95}{
    \begin{tabular}{lcc|cc|cc|ccc}
      \hline
       & \multicolumn{2}{c|}{LIBLINEAR} & \multicolumn{2}{c|}{LIBSVM/DC-SVM} & \multicolumn{2}{c|}{ML-SVM} 
       & \multicolumn{3}{c}{AML-SVM} \\ \cline{2-10}
            Datasets   &G-mean&C-Time  &G-mean  &C-Time  &G-mean     &C-Time  &G-mean    &C-Time    &W-Time    \\ \hline
            HIGGS      &0.54  &4406    & -      & -      &\bf{0.62}  &3283    &0.61      &\bf{2134} & \bf{882} \\
            SUSY       &0.68  &1300    & -      & -      &0.74       &1116    &\bf{0.76} &\bf{881}  & \bf{412} \\ \hline
    \end{tabular}
    }
  \end{center}
  \label{tab:aml_mlsvm_journal_paper_large_data_time_performance_measures}
  \vspace{-.5cm}
\end{table}

\subsection{Implementation}
The ML-SVM was developed using C++ and PETSc library~\cite{petsc-user-ref_short}. 
The following libraries are used for various parts of the framework. The FLANN library 
for finding approximate k-nearest neighbors before coarsening, and METIS library
for graph partitioning during the refinement. 
The LIBSVM~\cite{chang2011libsvm} as the underlying solver for SVM. The implementation is based on a single CPU core, and all the speedups are through algorithmic improvement without leveraging parallel processing. 
The proposed framework is developed based on the ML-SVM library. 
Moreover, the OpenMP library is speed up the parameter fitting at each level using multi-threading. 
Since the number of parameter fitting used for experiments is 9 in the first stage and 4 in the second stage of nested uniform design, we do not evaluate the speedups by increasing the number of threads. The coarsening implementation is sequential and based on a single CPU core.
All experiments for data sets with less than 1M data points have executed on a single machine with CPU Intel Xeon E5-2665 2.4 GHz and 64 GB RAM. For larger data sets, we used one single machine with CPU Intel Xeon E5-2680v3 2.5 GHz and 128 GB RAM.
%
%
%
\subsection{ML-SVM with AML-SVM Comparison}
The same experiment setup and hardware configuration is used for the results in this section and next section. Tables \ref{tab:aml_mlsvm_journal_paper_medium_data_time_performance_measures} and 
\ref{tab:aml_mlsvm_journal_paper_large_data_time_performance_measures} demonstrate a comparison across various libraries \cite{sadrfaridpour2019engineering}. The C-Time and W-Time are used to report CPU time and wall time respectively.
However, the purpose of experiments and the version of ML-SVM are different. The experiments are designed to have a better understanding of the multilevel framework and refinement process on each level of the hierarchical framework, while the earlier experiments are designed to compare the overall frameworks regardless of internal details. 
The CPU time is accurate metric for computational time of sequential programs such as LIBSVM, DC-SVM, and ML-SVM. Therefore, the CPU time is presented for them. However, The AML-SVM leverages multicore parallel processing which can reduce the wall time, hence, both CPU and wall times are reported.

In all experiments the data is normalized using z-score. 
The computational time reported in all experiments contains generating the $k$-NN graph. The computational time is reported in seconds unless it is explicitly mentioned otherwise.

In each class, a part of the data is assigned to be the test data using $k$-fold cross validation. We experimented with $k=5$ and $10$ (no significant difference was observed). 
The experiments are repeated $k$ times to cover all the data as test data. The data randomly shuffled for each $k$-fold cross validation.
The presented results are the averages of performance measures for all $k$  folds. 
Data points which are not in the test data are used as the training data in $\JJ^{+(-)}$. 
The test data is never used for any training or validation purposes. 
%
%

We run 18 experiments per dataset using nested cross-validation with k-fold where $k \in \{5,10\}$. with various configurations to evaluate the performance of the proposed method (AML). For each data set, we evaluated the combination of important parameters such as interpolation order (r), validation sample ratio (v), and where we stop to partition the data in ML method, which we define as stopping criteria for the AML method. We controlled all the other variables only to consider the effect of new method for an exactly similar configuration. Each combination of configuration provide slightly different results per dataset, and therefore, we evaluate 18 combinations of parameters and present the statistical information per dataset. The goal of the experiment is to have a well-designed experiment for comparing the current and new methods. We provide the related results from the MLSVM paper for convenient comparison of AML method with other states of the arts SVM solvers such as LIBSVM and DC-SVM. 

\begin{figure}
    \centering
    \includegraphics[width=\amlResultFig\linewidth]{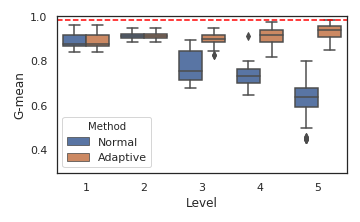} 
    
    \vspace{-0.5cm}
    
    \includegraphics[width=\amlResultFig\linewidth]{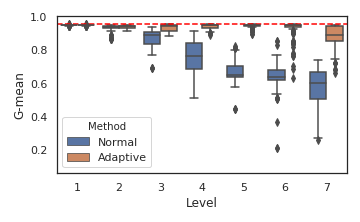} 	    

    \vspace{-0.5cm}
    
    \includegraphics[width=\amlResultFig\linewidth]{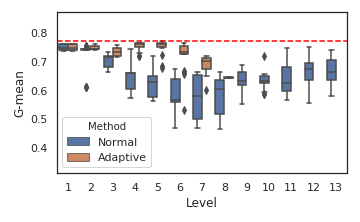} 	    
    
    \vspace{-0.6cm}
    
    \caption{Comparing the prediction quality across levels for Clean, Cod, and SUSY data set from top to bottom}
    \label{fig:aml_plot_clean_cod_susy}
\end{figure}

For each configuration, we present three sets of results over k-fold cross-validation: (1) Classification quality: we plot the statistics of G-mean at each level of uncoarsening; (2) Computational Performance: the average running time for both coarsening and uncoarsening framework is reported; and (3)  An overall comparison only for the final results of the whole framework without details for levels are provided in Tables~\ref{tab:aml_mlsvm_journal_paper_medium_data_time_performance_measures} to~\ref{tab:aml_tbl_time}. We only plot the statistics for Clean, Cod, and SUSY data set in figure \ref{fig:aml_plot_clean_cod_susy}. The rest of plots are available online at the GitHub repo.

\begin{table}[tb]
\caption{Computational Time (seconds)}
\label{tab:aml_tbl_time}
\resFontSize
\centering
\begin{tabular}{lc|cc}
    \hline
    &\multicolumn{1}{c}{ML-SVM}    &\multicolumn{2}{c}{AML-SVM}    \\ \cline{2-4} 
    Dataset        & CPU Time     & CPU Time  & Wall Time      \\ \hline
    Advertisement  &      67.6    &   126.0   & 23.84      \\
    Buzz           &      44.6    &   269.2   & 64.13      \\
    Clean          &       6.0    &    20.6   & 4.32       \\
    Cod-RNA        &       7.0    &    75.2   & 16.83      \\
    Forest         &    4993.3    &  2148.7   & 394.97     \\
    Letter         &       8.2    &    13.9   & 3.04       \\
    Ringnorm       &       3.1    &     3.1   & 0.77       \\
    Twonorm        &       0.8    &     1.4   & 0.42       \\
    HIGGS          &    6064.3    &  2134.4   & 882.17     \\
    SUSY           &     664.3    &   881.2   & 411.96     \\ \hline
\end{tabular}

\vspace{-.5cm}
\end{table}

The number of levels in coarsening and uncoarsening is the same, and we use them interchangeably for the rest of this section.
The Advertisement dataset has 3,279 data points and 1,558 features. It has a large number of features and a small number of data points. The number of levels in the coarsening phase, only depends on the number of data points. Therefore, there are only 3 levels in (un)coarsening phase for the Advertisement dataset. 
Both ML-SVM and AML-SVM frameworks have identical results for the first and second levels. However, the AML-SVM achieved higher quality on the final level. All the qualities are reported based on G-mean performance measures.

The Buzz dataset has 140,707 data points, 77 features, and 7 levels of coarsening. The quality of both methods is comparable except for a significant drop on the 2nd level for the ML-SVM quality. 
The Clean dataset has 6,598 data points, 166 features, and 5 levels of coarsening. The ML-SVM's quality is descending from the 3rd level to the 5th level. However, while the AML-SVM has not decreased, in the 4th and 5th levels, it achieves the highest quality across all the levels.
The Cod-RNA dataset is related to Breast Cancer. It has 59,535 data points and 8 features. It has a similar decreasing trend as the Clean dataset.

The Forest (Cover Type) dataset is one of the larger data sets with 581,012 data points and 54 features. Contrary to the earlier data sets, the quality (G-mean) at the first (coarsest) level is lower than most of the levels. The ML-SVM quality is increased in the 2nd level but continues to descend til level six. The quality at 7th level is increased with a slight decrease in the following levels. 
The AML-SVM recovers the descend at the third level and achieves the highest G-mean in third and fourth levels.

We designed early stopping for the AML-SVM to reduce the computational complexity. Therefore, as the size of training data reaches a threshold $\theta$ in Algorithm ~\ref{alg:aml_general_refinement}, the refinement process stops. 

The ML-SVM has a descending trend for quality on the Letter data set similar to Clean and Cod-RNA data sets. However, AML-SVM has no decrease in quality.
The Ringnorm dataset has 7,400 data points and 20 features. The ascending quality is observed for both methods without any drop in qualities.
The Twonorm dataset has the same size and number of features as the Ringnorm.
The ML-SVM quality drops in the 3rd and 4th levels. The adaptive methods achieve high-quality results on all levels.

The quality of ML-SVM for both HIGGS and SUSY data set has a similar trend of high quality at the coarsest levels following with lower quality in the middle levels. At the finer levels, there is a slight improvement, but the variance of qualities is larger than the coarser levels.
The adaptive method for the SUSY dataset achieves high quality up to level 6. In levels 7 and 8 the quality of adaptive method drops as well, and at level 8, the adaptive method stops. For the HIGGS dataset, the adaptive method has better quality compared to the normal method. At level 9, the adaptive method stops to reduce the computational complexity.

\section{Conclusion}

In this paper, we introduced a novel adaptive recovery technique for under/over-fitting challenges in hierarchical frameworks which performs well on nonlinear support vector machines. Our results based on 18 distinct configurations of essential parameters for both coarsening and uncoarsening phases show that the adaptive approach is less sensitive to changes in hyper-parameters. 
A significant advantage is improved computational time.
The adaptive method has reduced the prediction quality variance across various levels in the multilevel framework, which provides a more robust solution.
%

On massive data sets, the training size may increase significantly at the fine levels. Our proposed early stopping functionality improves the computational performance as it prevents computationally expensive fine level training. 
In addition, we developed multi-threading support for parameter fitting, which utilizes more processing power on single computing node and increases the overall performance. 

The proposed approach can be extended to a hierarchical learning framework with a clear objective for quality.
While our exhaustive results demonstrate the ability of classifiers to learn a high-quality model at the coarse and middle levels, the model selection over many models, or training an ensemble model can be studied further.
For large-scale problems, sampling is a traditional approach to reduce the computational time for intermediate tasks such as feature selection or dimensionality reduction. The proposed framework is a useful alternative for blind sampling of data. 




\bibliographystyle{IEEEtran}
\bibliography{full_bib}

\end{document}